# Trust-Based Incentive Mechanisms in Semi-Decentralized Federated Learning Systems

Ajay Kumar Shrestha [0000-0002-1081-7036]

Vancouver Island University, Nanaimo BC V9T1P5, Canada
ajay.shrestha@viu.ca

**Abstract.** In federated learning (FL), decentralized model training allows multiple participants to collaboratively improve a shared machine learning model without exchanging raw data. However, ensuring the integrity and reliability of the system is challenging due to the presence of potentially malicious or faulty nodes that can degrade the model's performance. This paper proposes a novel trust-based incentive mechanism designed to evaluate and reward the quality of contributions in FL systems. By dynamically assessing trust scores based on factors such as data quality, model accuracy, consistency, and contribution frequency, the system encourages honest participation and penalizes unreliable or malicious behavior. These trust scores form the basis of an incentive mechanism that rewards high-trust nodes with greater participation opportunities and penalties for low-trust participants. We further explore the integration of blockchain technology and smart contracts to automate the trust evaluation and incentive distribution processes, ensuring transparency and decentralization. Our proposed theoretical framework aims to create a more robust, fair, and transparent FL ecosystem, reducing the risks posed by untrustworthy participants.

**Keywords:** Federated Learning, Blockchain, Artificial Intelligence, Trust, Incentive, Smart Contracts.

## 1 Introduction

In recent years, federated learning (FL) has emerged as a privacy-preserving approach to machine learning, enabling collaborative model training across decentralized nodes without the need for centralizing raw data [1, 2]. By keeping sensitive information on local devices and only sharing model updates, FL has found applications in sectors where data privacy is of paramount importance, such as healthcare, finance, and the Internet of Things (IoT) [3, 4]. However, despite these advantages, the decentralized nature of FL introduces several challenges related to the integrity and trustworthiness of participating nodes [5, 6].

One of the central issues in FL is the vulnerability to malicious or faulty nodes [7, 8]. These participants can provide inaccurate or corrupted model updates, deliberately or unintentionally, which can significantly degrade the overall model performance [9, 10]. Additionally, some nodes may attempt to manipulate the learning process for personal gain, such as withholding data or submitting misleading updates to affect the final model [11, 12]. As the FL systems scale, ensuring the quality and reliability of each node's contributions becomes increasingly difficult, especially in environments where trust is limited [6, 13].

To address these challenges, this paper proposes a trust-based incentive mechanism for FL systems. It introduces a dynamic trust evaluation process that continuously assesses each node's contribution to the system based on multiple factors, including model accuracy, data quality, and update frequency. Nodes that consistently provide high-quality contributions will receive higher trust scores, leading to increased incentives such as access to more rounds of training or financial rewards. Conversely, nodes with low trust scores will be penalized through reduced participation or restricted access to shared resources. By establishing a system of rewards and penalties based on trust, we aim to mitigate the risk of model degradation due to malicious or faulty behavior. Unlike prior work that typically applies static or unidimensional metrics, our framework employs a real-time, multi-factor trust scoring system that directly governs participation and incentives, drawing inspiration from prior multi-factor incentive models in FL [14].

The proposed trust-based system is further strengthened by the integration of blockchain and InterPlanetary File System (IPFS) technologies. Blockchain's decentralized and immutable ledger records round metadata, IPFS content hash or content identifiers (CIDs) for the global model and report, and a compact digest of accepted contributors, payouts, and slashing events. Smart contracts execute payouts and slashing per this digest; trust computation and scoring remain off-chain. They facilitate reducing the need for central oversight and fostering a transparent and fair ecosystem for all participants [15]. Complementing this, IPFS is employed as a decentralized storage layer for model artifacts, enhancing auditability, verifiability, and system scalability by offloading data storage from the blockchain itself [6, 8, 16]. This integration differs from existing blockchain-FL approaches by tightly coupling trust dynamics with on-chain smart contract enforcement and decentralized model storage, enabling automated, auditable, and adaptive control of node behavior.

In this paper, we outline the theoretical model behind our trust-based incentive system and discuss its potential impact on improving the reliability, security, and scalability of FL. We also explore the challenges and trade-offs inherent in designing such a system, including computational overhead and the balance between rewards and penalties. Ultimately, our goal is to create a more resilient FL environment where nodes are motivated to contribute honestly and consistently, ensuring the success of decentralized machine learning initiatives. This work is distinct in its comprehensive treatment of trust as both a dynamic evaluative metric and an access-control mechanism specifically tailored for semi-decentralized federated learning (SDFL) architectures, a combination not previously addressed in the literature.

The remainder of the paper is structured as follows: Section II details the background and related work. The proposed trust-based incentive mechanism is presented in Section III, with the theoretical analysis following in Section IV. Finally, Section V concludes the paper.

## 2 Background And Related Work

FL has gained significant attention due to its ability to decentralize model training while preserving data privacy. In contrast to traditional machine learning approaches that rely on centralized data aggregation, FL enables multiple participants, or nodes, to train models locally on their own data and share only model updates with a central server or among peers. This decentralized framework addresses critical privacy concerns, making FL particularly attractive for industries such as healthcare, finance, and telecommunications, where data sensitivity is paramount [13].

Despite its advantages, FL faces significant challenges related to trust and data integrity. Since FL systems operate in decentralized environments, they are inherently vulnerable to malicious behavior from participating nodes. These behaviors can include data poisoning, where nodes intentionally submit faulty updates to degrade the global model, or free riding, where nodes benefit from the collective learning process without actively contributing quality data or model updates. These challenges underscore the importance of trust mechanisms that can evaluate the contributions of each participant and ensure the reliability of the shared model.

### 2.1 Mechanisms

Several approaches have been proposed to manage trust and incentivize honest behavior in FL. Traditional FL systems often rely on centralized aggregation mechanisms, where a trusted server aggregates model updates and filters out potentially harmful contributions [17–19]. However, centralized systems introduce a single point of failure and are vulnerable to attacks, thus contradicting the decentralized nature of FL [13]. Moreover, this approach assumes that the central server can be fully trusted, which may not always be the case.

To mitigate these risks, recent research has explored decentralized trust mechanisms that eliminate reliance on a single entity [16, 20]. These approaches often involve peer-to-peer trust evaluation among nodes, where participants assess each other's updates and contributions based on pre-defined metrics, such as model accuracy and contribution frequency [6, 21]. For instance, nodes that consistently provide accurate model updates may receive higher trust scores, while those that submit faulty or inconsistent updates may be penalized. While peer-to-peer trust evaluation reduces the need for centralized oversight, it also introduces challenges related to scalability and communication overhead, as trust must be computed and shared across a large network of nodes.

Incentive mechanisms are another critical component in promoting trustworthy behavior in FL systems [8]. Some models rely on monetary incentives, where participants are rewarded based on the quality of their contributions, often using blockchain or cryptocurrency-based systems. These financial incentives aim to encourage nodes to act honestly and contribute quality data and model updates [3]. However, improperly designed incentive structures can lead to unintended consequences, such as over-rewarding certain nodes or creating opportunities for manipulation. As a result, there is a need for dynamic and adaptive incentive mechanisms that align rewards with trust levels, penalizing malicious behavior while encouraging long-term honest participation [8, 10, 14].

### 2.2 Limitations of Existing Systems

Despite these developments, existing trust and incentive mechanisms face several limitations. One common challenge is the static nature of trust evaluation in many systems. Trust scores are often calculated based on a node's historical performance, but they fail to account for dynamic changes in behavior [3, 5, 7, 11, 12, 22]. For example, a node that was previously trustworthy may become malicious due to external factors, yet its trust score might not reflect this change until significant damage is done. Conversely, nodes that initially exhibit unreliable behavior may improve over time, but existing systems might take too long to recognize these improvements.

Moreover, most incentive mechanisms in FL are either too simplistic or too rigid [8, 13]. For instance, they often assign rewards solely based on a single metric, such as model accuracy, without considering other important factors like data quality, consistency, or the regularity of contributions. This lack of nuance can lead to suboptimal reward distribution, where nodes contributing high-quality updates in non-traditional ways, such as diverse datasets, may be undervalued, while nodes submitting frequent but low-quality updates are over-rewarded. Additionally, current incentive systems are often not integrated with trust evaluation, which limits their effectiveness in penalizing dishonest or malicious nodes.

### 2.3 Trust and Blockchain in Federated Learning

The integration of blockchain technology with FL has been proposed as a solution to several of these challenges. Blockchain's decentralized and immutable nature makes it a powerful tool for ensuring transparency and tamper-proof trust evaluation in FL systems [8, 23, 24]. By recording round metadata, IPFS CIDs for model/report artifacts, and payout/slashing events on a blockchain, participants can verify the legitimacy of the system and ensure that rewards and penalties are fairly enforced. Smart contracts, in particular, play a crucial role in automating the trust evaluation and incentive distribution processes, enabling real-time trust adjustments and preventing any single party from manipulating the system.

While blockchain-based systems offer significant potential for FL, they also introduce additional challenges, particularly related to computational overhead and scalability [2, 6]. Maintaining a blockchain requires considerable computational resources, and verifying transactions across a large network of nodes can introduce latency [7, 9]. Therefore, any integration of blockchain with FL must carefully balance trust transparency with system performance.

### 2.4 Gap in Literature and Proposed Solution

Despite significant progress in trust and incentive mechanisms for FL, existing systems fall short in supporting dynamic, multi-factor frameworks that can adapt to node behavior in real time. Current models are often rigid or overly simplistic, focusing on static trust metrics or single-factor incentive schemes that do not reflect the operational complexity of decentralized, multi-participant environments. While blockchain has emerged as a promising tool for enhancing transparency and security, its integration within FL remains relatively immature, with unresolved issues related to scalability, latency, and on-chain storage constraints limiting its practical adoption.

In response, this paper introduces a novel, fully decentralized trust-based incentive framework for FL that dynamically adjusts node trust scores and reward allocations based on multiple real-time performance indicators, including model accuracy, data quality, and contribution frequency. Unlike prior work, our system tightly couples these trust dynamics with smart contract enforcement to enable adaptive, auditable, and tamper-resistant control over collaborative behavior. Complementing this, IPFS is employed as a decentralized storage layer for model artifacts, which enhances auditability, verifiability, and system scalability by offloading data persistence from the blockchain itself. Collectively, these innovations enable a robust and extensible foundation for fair, efficient, and transparent collaboration in FL systems.

# 3 Trust-Based Incentive Mechanism

To address the challenges of malicious behavior and faulty nodes in FL, we propose a novel trust-based incentive mechanism that evaluates each node's contributions in real-time and adjusts their participation rewards accordingly. This mechanism builds upon existing trust evaluation frameworks but introduces dynamic, multi-faceted trust metrics and integrates them with a flexible incentive system. By leveraging blockchain, smart contracts and IPFS, the system ensures transparency and fairness in both trust evaluations and reward distributions, while maintaining the decentralized nature of FL.

## 3.1 Trust Evaluation Model

At the core of the proposed system is a dynamic trust evaluation model designed to assess the behavior and contributions of each node participating in the FL process. Unlike static trust systems that rely solely on historical data, our model continuously updates trust scores based on real-time performance metrics.

**Key Metrics for Trust Evaluation.** Trust in FL must be built on multiple dimensions to capture the full range of a node's contributions. The proposed trust evaluation model incorporates the following key metrics:

*Model Accuracy (A).* Each node's contribution is assessed based on the accuracy of its model updates. Nodes that submit model updates that align well with the global model's performance receive higher trust scores. This ensures that nodes are incentivized to provide accurate and relevant updates.

*Consistency (C).* Trust is not only built on individual contributions but also on the consistency of a node's performance over time. Nodes that consistently provide accurate and timely updates across multiple rounds of FL are rewarded with higher trust scores, whereas inconsistent nodes see their scores decline.

*Data Quality (D)*. Data quality is another crucial factor in trust evaluation. Nodes using high-quality, diverse datasets are more likely to produce better model updates, leading to higher trust scores. This can be inferred from the node's update history and its impact on the global model's performance.

*Update Frequency (U)*. The regularity with which a node participates in FL rounds is also considered. Nodes that contribute frequently and consistently add value to the system, while those that participate irregularly or intermittently are penalized. However, frequent low-quality updates will result in diminishing trust scores, ensuring that quality, not just quantity, is rewarded.

**Mathematical Model for Trust Calculation.** We define the trust score Ti of node i as a weighted sum of the aforementioned metrics:

$$T_i = \alpha A_i + \beta C_i + \gamma D_i + \delta U_i \tag{1}$$

- $A_i$: Accuracy of the node's model updates relative to the global model,
- $C_i$: Consistency in the node's contributions over multiple rounds,
- $D_i$: Data quality inferred from the node's contributions,
- $U_i$: Frequency of updates,
- $\alpha, \beta, \gamma, \delta$: Weights assigned to each metric, reflecting the importance of each factor in the overall trust evaluation.

These weights can be adjusted based on specific requirements of the FL environment or application. For instance, systems where model accuracy is critical might assign a higher value to $\alpha$, whereas systems prioritizing participation might increase $\delta$. The trust score $T_i$ is updated after every FL round, allowing for dynamic trust adjustment based on real-time performance.

**Trust Decay and Trust Recovery.** To prevent stagnation in trust scores and foster long-term accountability, the proposed system includes mechanisms for trust decay and trust recovery:

*Trust Decay*. When a node remains inactive in the FL process for a prolonged period, its trust score gradually decreases. This prevents dormant or disengaged nodes from retaining high trust levels indefinitely and ensures that trust reflects recent participation. This decay is modeled using an exponential function:

$$T_i(t) = T_i(t0) \cdot e^{-\lambda(t-t0)} \tag{2}$$

Where $T_i(t)$ denotes the trust score of node $i$ at time t, *t0 is the time of last active participation, and* $\lambda$ is the decay constant controlling the rate of trust reduction. This mechanism discourages passive or intermittently engaged participants from indefinitely retaining elevated trust.

*Trust Recovery*. Conversely, nodes that initially exhibit low-quality or inconsistent behavior but subsequently improve their contributions can gradually regain trust over time. This mechanism encourages positive behavioral change and supports long-term participation by offering a path to reputational redemption. The recovery follows a bounded growth function:

$$T_i(t+1)=T_i(t)+\eta\cdot(T_{max}-T_i(t))\cdot\Delta_i \quad (3)$$

Where $\eta$ is the recovery rate, $T_{max}$ is the maximum attainable trust score, and $\Delta_i$ captures a normalized measure of recent improvement (e.g., relative accuracy gains or consistency of updates). This enables the system to reward behavioral correction and fosters long-term engagement.

These dynamic trust mechanisms are inspired by time-frame-based trust models in peer-to-peer systems, time-domain trust decay frameworks, and dynamic evaluation mechanisms in blockchain-enabled IoT networks [18, 19, 25]. Integrated into our protocol, these temporal components help sustain fairness, responsiveness, and long-term participation in FL environments. The operational assumptions follow the server (or cluster coordinator) computing per-round metrics using the submitted updates and a held-out validation set or proxy evaluation. Raw data never leaves the nodes. All metric values are normalized to [0, 1], and the trust score $T_i$ is updated each round using the existing formula (1) and the decay/recovery rules (2) and (3) already defined. The default smoothing/decay coefficients and weights are deployment parameters and can be tuned per task.

### 3.2 Incentive Mechanism

The proposed trust-based incentive mechanism uses each node's trust score to determine its level of participation in the system, as well as its potential rewards. The goal is to incentivize honest and high-quality contributions while penalizing malicious behavior or negligence.

**Participation and Rewards are Determined by Policy.** The incentive mechanism uses $T_i$ to make deterministic decisions about admission, throttling, suspension, rewards, and penalties. Node admission, throttling, and suspension follow Table 1 (thresholds $\tau_{\text{admit}}, \tau_{\text{prob}}$; cadence limits $R$; suspension window $H$). When capacity is limited to $K_{\max}$ participants, the coordinator forms an eligibility set $E$ consisting of all active nodes plus a throttled subset of probation nodes (Table 1). $E$ is sorted by descending $T_i$; ties break by higher $C_i$, then by fewer recent strikes. The top min $(K_{\max}, |E|)$ nodes are admitted. The admission list is referenced in the round header, and its hash is committed on-chain during finalization. Nodes on probation are admitted at most once every $R$ rounds until $T_i \geq \tau_{\text{admit}}$. Nodes with $T_i < \tau_{\text{prob}}$ are suspended for $H$ rounds and are not considered for admission during lockout; after $H$, they move to Probation and may re-enter via recovery. Screening and aggregation follow Table 2 (accept iff non-negative accuracy gain and $D_i \geq \tau_D$; trust-weighted robust aggregation). Only accepted updates influence the global model (accepted set and weights are listed in the round report CID). Rewards and slashing follow Table 3 (budget $B_t$ split by utility with probation cap $\phi$; stake-based strikes $S$ within window $W$ trigger slashing $\mu$ and suspension $H$; temporary cap $T_{\text{cap}}$during rehabilitation $W_{\text{rehab}}$).

**Table 1.** Policy Box – Admission, Probation, Suspension.

| Given a node's current trust $T_i$: |
|---|
| Admit to next round if $T_i \geq \tau_{\text{admit}}$ |
| Probation if $\tau_{\text{prob}} \leq T_i < \tau_{\text{admit}}$: admitted at most once every $R$rounds. |
| Suspend for $H$rounds if $T_i < \tau_{\text{prob}}$. |
| *Reference profile:* $\tau_{\text{admit}} = 0.40$, $\tau_{\text{prob}} = 0.25$, $R = 2$, $H = 2$. |

**Table 2.** Policy Box – Screening and Aggregation.

| An update is accepted for aggregation iff: |
|---|
| (i) accuracy gain is non-negative |
| (ii) update-quality (similarity) exceeds a threshold $D_i \geq \tau_D$. |
| Accepted updates are aggregated with **trust-weighted robust aggregation** (e.g., median/trim-mean with weights proportional to $T_i$). |
| *Reference profile:* $\tau_D = 0.20$. |

**Table 3.** Policy Box – Rewards and Penalties.

| Title | Description |
|---|---|
| Rewards | A round budget $B_t$ is distributed only among accepted updates, in proportion to a utility score combining (non-negative) accuracy gain and trust weight. Probationary nodes receive at most a fraction $\phi$ of their computed share. |
| Penalties | Each node escrows a small stake at registration. If an update fails screening, a strike is recorded; upon $S$ strikes within a window of $W$ rounds, the contract slashes a fraction $\mu$ of the stake and the node is suspended $H$ rounds. After any slashing, trust is temporarily capped at $T_{\text{cap}}$ during a rehabilitation window of $W_{\text{rehab}}$ rounds. |
| *Reference profile:* $\phi = 0.5$, $S = 2$within $W = 5$, $\mu = 0.10$, $T_{\text{cap}} = 0.60$, $W_{\text{rehab}} = 5$. | |

### 3.3 Blockchain and Smart Contract Integration

To ensure fairness and transparency in the trust evaluation and incentive distribution processes, we propose the integration of blockchain technology. Blockchain provides a decentralized and immutable platform for recording round metadata and enforcing incentive mechanisms, while detailed trust and metrics remain off-chain.

**Smart Contracts for Automated Incentives**. Smart contracts, deployed on a blockchain, play a crucial role in automating the trust-based incentive system. They record the final round state and enforce payouts and slashing based on the coordinator's digest:

*Record Round State.* Persist (roundId, cid_model, cid_report, digest), where digest (e.g., a Merkle root) summarizes accepted contributors, computed payouts, and any slashing events computed off-chain.

*Distribute Rewards*. Transfer token payouts per digest.

*Apply Slashing*. Execute slashing and suspensions per policy (strike windows) as specified in the digest.

*Immutability and Transparency*. On-chain records reference IPFS artifacts, providing a verifiable trail of decisions and outcomes. This builds trust among participants by making enforcement auditable.

**Decentralized Storage with IPFS**. To reinforce blockchain's transaction immutability, the system integrates IPFS as a decentralized storage layer for off-chain artifacts such as model snapshots and round reports (metrics, accepted set, weights, payouts, slashing intents). Each file is referenced on-chain by its CID, ensuring data integrity and tamper resistance while keeping chain storage minimal. This approach reduces blockchain storage overhead and enables efficient sharing of large, FL artifacts across participants in a scalable and verifiable manner.

**Table 4.** Concrete Interface (on-chain).

| Procedure | Purpose |
|---|---|
| registerNode (pubkey, stake | Escrow a node's stake. |
| finalizeRound (roundId, cid_model, cid_report, digest) | Record IPFS CIDs and a compact digest (e.g., Merkle root) of accepted nodes, rewards, and any slashing events. |
| distributeRewards (roundId) | Transfer rewards to accepted nodes. |
| slash (nodeId, fraction, reason) | Executed within finalizeRound if policy conditions are met. |

**Off-chain Coordinator.** For the off-chain, the coordinator executes the admission/screening/aggregation, and reward policies; publishes the global model and a per-round report to IPFS; obtains their CIDs; and calls finalizeRound. Only CIDs and digests are on-chain; trust scores and detailed metrics remain off-chain but are verifiable via the IPFS report.

### 3.4 Trust-Based Leadership and Aggregation

In cluster-based FL environments, the trust score also plays a pivotal role in determining leadership roles. The cluster aggregator role is assigned to the highest-trust node in the cluster at the start of each round; ties are broken by higher consistency $C_i$. The aggregator is responsible for combining local model updates from other nodes. During aggregation, only accepted updates (per Table 2) are included, and contribution weights are monotone in trust, with nodes having higher trust scores weighted more heavily so their contributions have greater impact on the global model.

### 3.5 Workflow Illustration of the Semi-Decentralized FL Protocol

To further clarify the interactions among system components, Fig. 1 presents a UML sequence diagram of a single training round in the proposed semi-decentralized architecture, highlighting the separation of concerns between off-chain coordination and on-chain enforcement, and the use of IPFS for verifiable artifact storage. The Requester/Coordinator (R) executes model distribution, screening, trust updates, and aggregation off-chain; the Smart Contract (S) records round metadata, enforces penalties, and distributes rewards on-chain; IPFS provides content-addressed storage for the global model and the round report; and workers (W1-W3) perform local training and submit updates asynchronously.

**Round Header (On-chain Pointer).** The round begins with publish_round_header, where R records the round ID, thresholds, and hashes of the admission list and state partitions (Active/Probation/Suspended), along with prior CIDs. This establishes an auditable reference for who is eligible to participate and under what rules.

**Model Distribution (Off-chain)**. R disseminates the current global model and round header to the admitted set. This reflects the priority policy (high-trust first) and restriction policy (probation throttling and temporary suspensions) defined earlier.

**Local Training and Asynchronous Submissions.** Each admitted worker performs local_train() and sends submit_update($\Delta$w, qualitySignal) to R. Any optional quality signal is treated as a hint; authoritative screening is performed by R.

**Screening and Trust Update (Off-chain).** For each received update, R applies the deterministic screens: non-negative accuracy gain and update-quality at or above $\tau_D$. Accepted updates trigger a per-node trust update; rejected updates are recorded as strikes to be considered at finalization. No slashing occurs mid-round.

**Trust-Weighted Robust Aggregation (Off-chain)**. R aggregates only the accepted updates using a robust method with weights monotone in trust, then updates the global model. This step operationalizes "priority access" during aggregation while preserving robustness to outliers.

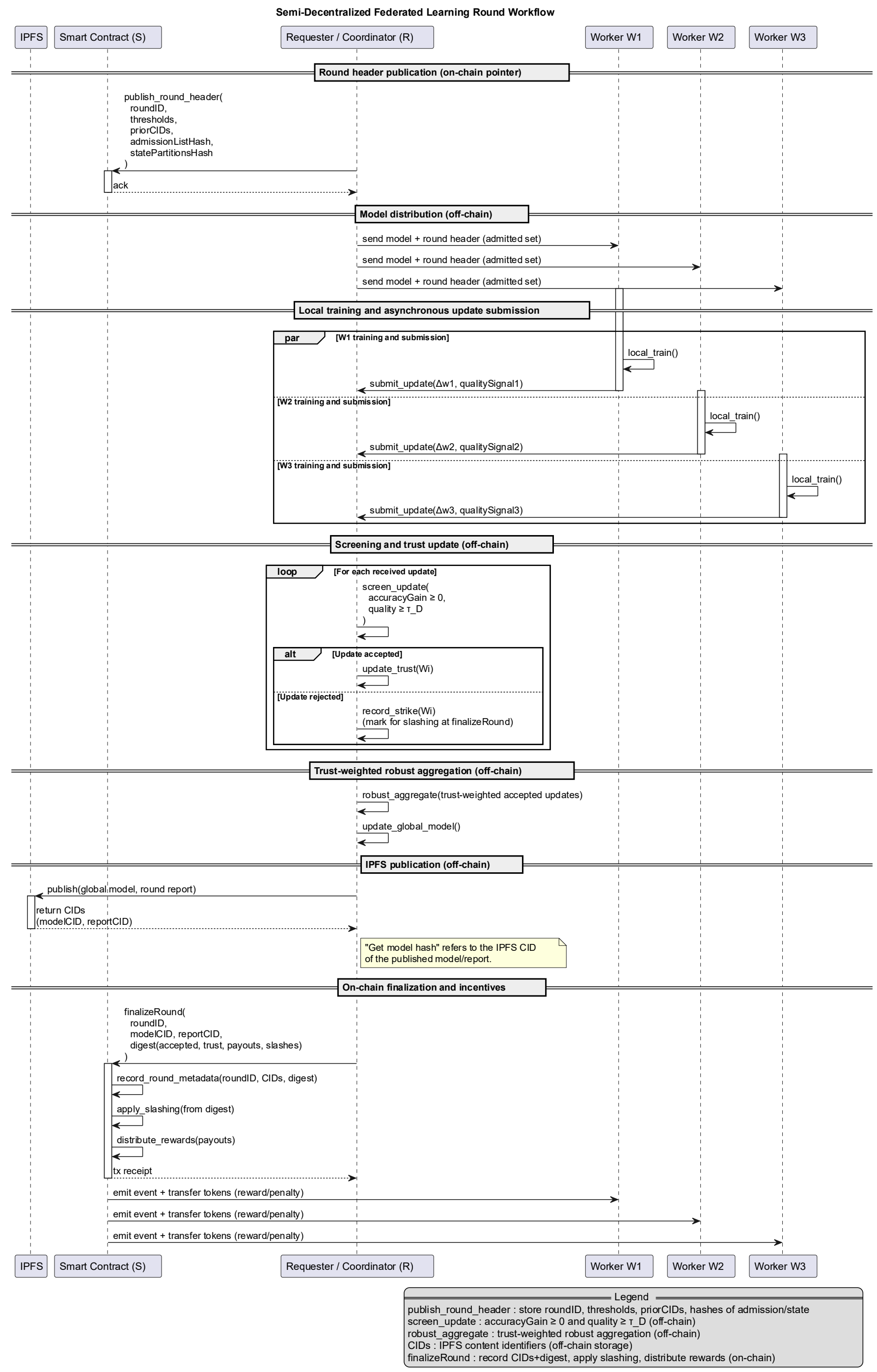

**Fig. 1.** UML sequence diagram of the semi-decentralized federated learning round.

**IPFS Publication (Off-chain)**. R publishes the new global model and a compact round report (accepted set, trust values, weights, rewards, and any slashing intents) to IPFS, obtaining CIDs (modelCID, reportCID). "Get model hash" in Fig. 1 denotes retrieving these CIDs.

**On-chain Finalization and Incentives**. R calls finalizeRound(roundID, modelCID, reportCID, digest), where digest (e.g., a Merkle root) summarizes accepted contributors, computed payouts, and pending slashes. S records the CIDs and digest, applies slashing according to the policy (e.g., repeated failed screens), and executes distribute_rewards. The contract emits events and transfers tokens, providing a tamper-resistant, auditable incentive trail.

This round-oriented workflow makes explicit how behind earlier policy statements: high-trust nodes receive deterministic priority via the admitted set and trust-weighted aggregation; low-trust nodes are restricted through probation throttling and temporary suspension reflected in the header hashes and enforced at selection time; and all incentives and penalties are verifiable through on-chain records that point to IPFS-hosted artifacts. The same pattern repeats each round, with updated trust and state partitions informing the next publish_round_header. This architectural flow builds upon earlier work [6], integrating asynchronous trust-penalized updates in blockchain-enabled FL systems.

# 4 Theoretical Analysis

The proposed trust-based incentive mechanism aims to enhance the robustness and reliability of FL systems by addressing key challenges such as malicious behavior, model degradation, and inconsistent contributions. In this section, we analyze the theoretical benefits, trade-offs, and potential challenges of integrating trust-based incentives into FL environments.

Under Table 2, only updates passing (gain $\geq 0 \wedge D_i \geq \tau_D$) influence aggregation; with trust-weighted robust aggregation, low-trust or marginal updates have a bounded impact. Under Table 3, repeated failed screens ($\geq S$ within $W$ rounds) deterministically trigger slashing $\mu$ and suspension $H$, limiting the cumulative influence of unreliable nodes across rounds. Table 1's thresholds ($\tau_{\text{admit}}, \tau_{\text{prob}}$) and cadence $R$ enforce deterministic priority for high-trust nodes and throttling for probationary ones, operationalizing "priority access" and "restriction" without ambiguity. The decay/recovery rules (2)-(3) further bound how long stale trust persists and how quickly improvement is recognized.

## 4.1 Impact of Trust Evaluation on Federated Learning

The trust evaluation model is designed to incentivize nodes to consistently contribute high-quality updates while discouraging malicious or unreliable behavior. The dynamic nature of the trust system ensures that nodes are continuously evaluated based on their performance, fostering a competitive yet cooperative environment.

**System Robustness and Security.** By introducing trust scores that are updated in real time, the proposed system enhances the robustness of FL models. Nodes that submit low-quality or malicious updates will quickly see their trust scores drop, limiting their participation and influence over the global model. This reduces the risk of data poisoning attacks, where faulty nodes intentionally corrupt the learning process by submitting misleading or erroneous updates.

Furthermore, the inclusion of penalties, such as reduced participation or financial loss, creates a deterrent for nodes that might otherwise attempt to disrupt the system. Honest nodes are rewarded for their consistent contributions, while malicious behavior is discouraged through automatic sanctions. As a result, the overall security of the FL process is strengthened, as nodes are more likely to act in good faith to preserve their reputation and trust scores.

**Faulty Node Identification.** The trust evaluation model also serves as an efficient method for identifying faulty or unreliable nodes. As each node's trust score is updated after every round of FL, the system quickly detects nodes that consistently submit low-quality updates. These nodes are flagged by their declining trust scores, allowing the system to take corrective actions, such as reducing their participation or applying penalties.

Unlike traditional systems, where faulty nodes may continue to participate until they cause significant model degradation, the proposed trust-based system offers early detection of unreliable behavior. This allows the FL model to maintain its integrity and accuracy, even in the presence of adversarial or low-quality participants.

**Fairness in Participation.** One of the key advantages of the trust-based mechanism is its ability to maintain fairness in participation and reward distribution. Nodes that contribute meaningfully are rewarded not only with higher trust scores but also with increased participation opportunities and financial incentives. This ensures that the most reliable and valuable participants are prioritized, creating a merit-based system where honest behavior is consistently rewarded.

At the same time, the dynamic nature of trust scores allows nodes with initially low performance to improve over time. This ensures that the system remains inclusive, as underperforming nodes are given opportunities to regain trust through consistent and reliable participation, rather than being permanently penalized for early mistakes.

### 4.2 Trust-Based Incentives and Decentralization

The proposed incentive mechanism is designed to operate in decentralized environments, such as those enabled by blockchain and FL. One of the primary benefits of this approach is its ability to scale effectively across large, distributed networks without the need for centralized control.

**Scalability and Efficiency.** In traditional FL systems, scaling the network introduces challenges related to trust management and coordination among nodes. With the integration of blockchain, trust scores and incentive mechanisms are managed in a decentralized manner through smart contracts. This eliminates the need for a central authority to oversee trust evaluations and rewards, making the system more scalable and reducing potential bottlenecks.

Moreover, the dynamic nature of the trust-based incentives ensures that only high-trust nodes contribute significantly to the learning process. This optimizes resource allocation by prioritizing nodes that provide the most value, leading to faster model convergence and improved efficiency in large-scale FL systems.

**Transparency and Trust in Decentralized Systems.** By leveraging blockchain technology, the proposed system ensures transparency and immutability in trust evaluations and incentive distributions. All on-chain transactions, round metadata, IPFS CIDs, and reward/slashing events are recorded on a decentralized ledger that is accessible to all participants. This transparency fosters trust among nodes, as they can verify that the system operates fairly and rewards are distributed according to predefined rules.

In traditional systems, participants may have doubts about the fairness of reward distribution or trust evaluations, particularly if these processes are managed by a central entity. In contrast, blockchain's decentralized and tamper-proof nature ensures that no single participant or entity can manipulate the system, further reinforcing the integrity of the FL process.

### 4.3 Challenges and Trade-offs

While the proposed trust-based incentive mechanism offers significant advantages, there are several challenges and trade-offs that must be considered in its implementation.

**Computational Overhead.** One of the primary challenges is the computational overhead associated with trust evaluation and blockchain integration. Calculating trust scores for each node after every round of FL requires additional computational resources, particularly when multiple metrics (accuracy, consistency, data quality, etc.) are involved. Moreover, maintaining a blockchain ledger introduces additional transactional costs, as each trust score update and reward distribution must be recorded on the blockchain.

To mitigate these challenges, optimization techniques such as periodic trust score updates (instead of real-time updates) or the use of off-chain solutions for recording less critical transactions could be employed. These strategies help balance the benefits of transparency and security with the need for computational efficiency.

**Consistency and Convergence.** Another challenge lies in ensuring consistency and convergence in FL when some nodes are penalized or restricted from participation.

While low-trust nodes are penalized to maintain the model's integrity, this may result in fewer nodes contributing to the global model, potentially slowing down convergence.

To address this, the system must carefully balance the penalties imposed on low-trust nodes with the need for consistent participation from a diverse pool of nodes. Ensuring that nodes have opportunities for trust recovery will help maintain a broad base of participants, ultimately supporting faster convergence and higher model accuracy.

**Trade-offs in Incentive Design.** Designing the incentive structure requires careful consideration to avoid over-penalizing low-trust nodes or creating perverse incentives that encourage short-term behavior over long-term engagement. For instance, if penalties for low-trust nodes are too harsh, they may be discouraged from improving their behavior, leading to a drop in overall participation.

To strike the right balance, the system must implement adaptive incentives that reward honest behavior without completely excluding low-trust nodes. This could involve introducing gradual penalties or allowing nodes to recover trust more quickly based on significant improvements in performance over a set number of rounds.

**Complexity and Cost.** Per-round, trust metric computation is $O(|\mathcal{A}_t|)$ on the admitted set, and screening plus robust aggregation (median/trim-mean) are $O(|\mathcal{A}_t| \log |\mathcal{A}_t|)$. The on-chain *finalizeRound* commits constant-size CIDs and a digest independent of model size so gas cost scales with the number of payouts/slashes events rather than artifact sizes. In practice, runtime is dominated by the coordinator's evaluation/aggregation; large artifacts remain off-chain in IPFS.

## 5 Conclusion

In this paper, we propose a novel trust-based incentive mechanism for FL systems that aims to enhance the reliability and robustness of decentralized model training by evaluating the trustworthiness of nodes based on multiple dynamic metrics. Our model aims to address critical challenges related to malicious behavior, faulty nodes, and data quality by continuously assessing trust scores and aligning incentives to promote high-quality contributions. Through the integration of blockchain technology and smart contracts, we further ensured the transparency and fairness of trust evaluations and incentive distributions, making the system tamper-resistant and scalable. The proposed framework creates a fair and merit-based system where trustworthy nodes are rewarded, while unreliable or malicious nodes face appropriate penalties. By dynamically adjusting trust scores based on real-time performance, our system reduces the risk of model degradation caused by adversarial behavior and encourages consistent, high-quality participation from all nodes. This contributes to improved model accuracy, faster convergence, and greater system resilience. Despite its benefits, the proposed mechanism presents several challenges, such as computational overhead, the need to balance incentive design, and ensuring convergence when penalizing low-trust nodes. Future work will focus on optimizing the computational efficiency of trust evaluation, exploring off-chain solutions for transaction management, and developing more adaptive penalty and reward structures that strike a balance between discouraging malicious behavior and promoting long-term engagement.

## References

1. Qu, Y., Xu, C., Gao, L., Xiang, Y., Yu, S.: FL-SEC: Privacy-preserving decentralized federated learning using SignSGD for the Internet of Artificially Intelligent Things. IEEE Internet of Things Magazine. 5, 85–90 (2022). https://doi.org/10.1109/IOTM.001.2100173.
2. Singh, S., Rathore, S., Alfarraj, O., Tolba, A., Yoon, B.: A framework for privacy-preservation of IoT healthcare data using federated learning and blockchain technology. Future Generation Computer Systems. 129, 380–388 (2022). https://doi.org/10.1016/J.FUTURE.2021.11.028.
3. Said, N. Al: Federated learning for privacy-preserving AI: Challenges, applications, and future directions. Panamerican Mathematical Journal. 35, 358–368 (2025). https://doi.org/10.52783/PMJ.V35.I3S.4054.
4. Al Jasem, M.S., De Clark, T., Shrestha, A.K.: Toward decentralized intelligence: A systematic literature review of blockchain-enabled AI systems. Information (Switzerland). 16, 765 (2025). https://doi.org/10.3390/INFO16090765/S1.
5. Zhang, K., Song, X., Zhang, C., Yu, S.: Challenges and future directions of secure federated learning: a survey. Front Comput Sci. 16, 165817 (2022). https://doi.org/10.1007/s11704-021-0598-z.
6. Shrestha, A.K., Khan, F.A., Shaikh, M.A., Jaberzadeh, A., Geng, J.: Enhancing scalability and reliability in semi-decentralized federated learning with blockchain: Trust penalization and asynchronous functionality. 2023 IEEE 14th Annual Ubiquitous Computing, Electronics and Mobile Communication Conference, UEMCON 2023. 230–236 (2023). https://doi.org/10.1109/UEMCON59035.2023.10316006.
7. Martinez Beltran, E.T., Perez, M.Q., Sanchez, P.M.S., Lopez Bernal, S., Bovet, G., Perez, M.G., Perez, G.M., Celdran, A.H.: Decentralized federated learning: Fundamentals, state of the art, frameworks, trends, and challenges. IEEE Communications Surveys and Tutorials. 25, 2983–3013 (2023). https://doi.org/10.1109/COMST.2023.3315746.
8. Jaberzadeh, A., Shrestha, A.K., Khan, F.A., Shaikh, M.A., Dave, B., Geng, J.: Blockchain-based federated learning: Incentivizing data sharing and penalizing dishonest behavior. In: 5th International Congress on Blockchain and Applications (BLOCKCHAIN'23). pp. 186–195 (2023). https://doi.org/10.1007/978-3-031-45155-3_19.
9. Geren, C., Board, A., Dagher, G.G., Andersen, T., Zhuang, J.: Blockchain for large language model security and safety: A holistic survey. ACM SIGKDD Explorations Newsletter. 26, 1–20 (2025). https://doi.org/10.1145/3715073.3715075;WGROUP:STRING:ACM.
10. Kalapaaking, A.P., Khalil, I., Yi, X.: Blockchain-based federated learning with SMPC model verification against poisoning attack for healthcare systems. IEEE Trans Emerg Top Comput. 12, 269–280 (2024). https://doi.org/10.1109/TETC.2023.3268186.
11. Zhao, J., Bagchi, S., Avestimehr, S., Chan, K., Chaterji, S., Dimitriadis, D., Li, J., Li, N., Nourian, A., Roth, H.: The federation strikes back: A survey of federated learning privacy attacks, defenses, applications, and policy landscape. ACM Comput Surv. 57, 1–37 (2025). https://doi.org/10.1145/3724113.
12. Zhang, C., Yang, S., Mao, L., Ning, H.: Anomaly detection and defense techniques in federated learning: a comprehensive review. Artif Intell Rev. 57, 1–34 (2024). https://doi.org/10.1007/S10462-024-10796-1/FIGURES/9.

13. Tariq, A., Serhani, M.A., Sallabi, F.M., Barka, E.S., Qayyum, T., Khater, H.M., Shuaib, K.A.: Trustworthy federated learning: A comprehensive review, architecture, key challenges, and future research prospects. IEEE Open Journal of the Communications Society. 5, 4920–4998 (2024). https://doi.org/10.1109/OJCOMS.2024.3438264.
14. K.M., S., Nicolazzo, S., Arazzi, M., Nocera, A., Rafidha, R.R., P., V., Conti, M.: Privacy-preserving in blockchain-based federated learning systems. Comput Commun. 222, 38–67 (2024). https://doi.org/10.1016/j.comcom.2024.04.024.
15. Cassano, L., D'Abramo, J., Munir, S., Ferretti, S.: Trust and resilience in federated learning through smart contracts enabled decentralized systems. Proceedings - 2024 IEEE International Conference on Blockchain, Blockchain 2024. 663–668 (2024). https://doi.org/10.1109/BLOCKCHAIN62396.2024.00097.
16. Yang, J., Zhang, W., Guo, Z., Gao, Z.: TrustDFL: A blockchain-based verifiable and trusty decentralized federated learning framework. Electronics 2024, Vol. 13, Page 86. 13, 86 (2023). https://doi.org/10.3390/ELECTRONICS13010086.
17. Sánchez Sánchez, P.M., Huertas Celdrán, A., Xie, N., Bovet, G., Martínez Pérez, G., Stiller, B.: FederatedTrust: A solution for trustworthy federated learning. Future Generation Computer Systems. 152, 83–98 (2024). https://doi.org/10.1016/j.future.2023.10.013.
18. Zhou, R., Hwang, K.: PowerTrust: A robust and scalable reputation system for trusted peer-to-peer computing. IEEE Transactions on Parallel and Distributed Systems. 18, 460–473 (2007). https://doi.org/10.1109/TPDS.2007.1021.
19. Li, D., Zhu, Z., Cheng, C., Du, F.: TD-Trust: A time domain based p2p trust model. Lecture Notes in Computer Science (including subseries Lecture Notes in Artificial Intelligence and Lecture Notes in Bioinformatics). 7002 LNAI, 467–474 (2011). https://doi.org/10.1007/978-3-642-23881-9_61.
20. Shrestha, A.K.: Designing incentives enabled decentralized user data sharing framework, https://harvest.usask.ca/handle/10388/13739, (2021).
21. Wink, T., Nochta, Z.: An approach for peer-to-peer federated learning. Proceedings - 51st Annual IEEE/IFIP International Conference on Dependable Systems and Networks Workshops, DSN-W 2021. 150–157 (2021). https://doi.org/10.1109/DSN-W52860.2021.00034.
22. Liu, K., Yan, Z., Liang, X., Kantola, R., Hu, C.: A survey on blockchain-enabled federated learning and its prospects with digital twin. Digital Communications and Networks. (2022). https://doi.org/10.1016/J.DCAN.2022.08.001.
23. O'Hara, K.: Data trusts ethics, architecture and governance for trustworthy data stewardship. (2019). https://doi.org/10.5258/SOTON/WSI-WP001.
24. Delacroix, S., Lawrence, N.D.: Bottom-up data trusts: Disturbing the 'one size fits all' approach to data governance. International Data Privacy Law. 9, 236–252 (2019). https://doi.org/10.1093/IDPL/IPZ014.
25. Tu, Z., Zhou, H., Li, K., Song, H., Yang, Y.: A blockchain-based trust and reputation model with dynamic evaluation mechanism for IoT. Computer Networks. 218, 109404 (2022). https://doi.org/10.1016/J.COMNET.2022.109404.